\documentclass{article} 
\usepackage{iclr2016_conference,times}
\usepackage{url}
\usepackage[pdftex]{graphicx}
\usepackage{algorithm,algorithmic}
\usepackage{hyperref}

\title{Improving Back-Propagation by Adding an \\
Adversarial Gradient}

\author{
Arild N\o{}kland \\
Trondheim, Norway \\
\texttt{arild.nokland@gmail.com}
}

\begin{document}

\maketitle

\begin{abstract}
The back-propagation algorithm is widely used for learning in artificial neural networks. A challenge in machine learning is to create models that generalize to new data samples not seen in the training data. Recently, a common flaw in several machine learning algorithms was discovered: small perturbations added to the input data lead to consistent misclassification of data samples. Samples that easily mislead the model are called adversarial examples. Training a ''maxout'' network on adversarial examples has shown to decrease this vulnerability, but also increase classification performance. This paper shows that adversarial training has a regularizing effect also in networks with logistic, hyperbolic tangent and rectified linear units. A simple extension to the back-propagation method is proposed, that adds an adversarial gradient to the training. The extension requires an additional forward and backward pass to calculate a modified input sample, or mini batch, used as input for standard back-propagation learning. The first experimental results on MNIST show that the ''adversarial back-propagation'' method increases the resistance to adversarial examples and boosts the classification performance. The extension reduces the classification error on the permutation invariant MNIST from 1.60\% to 0.95\% in a logistic network, and from 1.40\% to 0.78\% in a network with rectified linear units. Results on CIFAR-10 indicate that the method has a regularizing effect similar to dropout in fully connected networks. Based on these promising results, adversarial back-propagation is proposed as a stand-alone regularizing method that should be further investigated.

\end{abstract}

\section{Introduction}
Supervised feed-forward neural networks are often trained using the back-propagation learning algorithm (\citet{Hinton86}) and stochastic gradient descent. This is a powerful method able to train networks with hidden layers and learn complex mappings between input and output neurons. However, the usefulness of a model depends not only on the ability to learn the training data, but also the ability to classify new data not seen in the training data set. The property of generalization is important especially if training data is limited. Standard back-propagation tends to learn the training data too well when trained for a long time. To avoid this, early stopping is often used to obtain the best results on the test data. 

Several regularization methods have been used with back-propagation to increase generalization. Popular methods include pre-training (\citet{HintonOT06}, \citet{SalakhutdinovH09}, \citet{VincentLBM08}), dropout (\citet{SrivastavaHKSS14}), noise injection (\citet{Matsuoka92}), weight decay, and recently, batch normalization (\citet{IoffeS15}). 

After the discovery of adversarial examples (\citet{SzegedyZSBEGF13}), methods for understanding the flaw and increasing the resistance to these examples has been explored (\citet{NguyenYC14}, \citet{GuR14}, \citet{GoodfellowSS14}, \citet{FawziFF15}). \citet{GoodfellowSS14} showed that augmenting the training data set with adversarial examples made the model more resistant against adversarial perturbations. Additionally, adversarial training regularized the model and improved the classification performance in a ''maxout'' network model on the MNIST data set. \citet{MiyatoMKNI15} showed that adding a label-independent adversarial objective to the training, together with batch normalization, gave better results on MNIST than any method using dropout.

If adding an adversarial objective improves the generalization property of the network, adversarial training might be used as a regularization method on it's own, without dropout or batch-normalization. The aim of this paper is to show through experiments that adversarial training regularizes the model, and that no data augmentation is required.

\section{Method}
One way to generate adversarial examples is to use the ''fast gradient sign method'' (\citet{GoodfellowSS14}). A perturbation is added to the original data sample, and this perturbation is proportional to the sign of the gradient back-propagated from the output to the input layer.

The adversarial back-propagation method uses the fast gradient sign method and adds a forward and backward pass, in order to calculate the gradient. The perturbated sample is used for learning. This can be seen as a way to make use of the gradient at the input layer. This gradient is not used in standard back-propagation, even though it is easily available by an additional gradient back-propagation step from first hidden layer to the input. 

The objective function for adversarial back-propagation is:

\begin{equation}
\label{eq:1}
F(\theta, x, y) = J(\theta, x + \mu, y)
\end{equation}
\begin{equation}
\label{eq:2}
\mu = \epsilon \ sign (\nabla_xJ(\theta, x, y))
\end{equation}

where $J(\theta, x, y)$ is an objective function like squared error or cross-entropy loss, $\theta$ is the network weights, $x$ is the input, $y$ is the desired output, $\mu$ is the fast gradient sign perturbation, and $\epsilon$ is the magnitude of the perturbation. $\nabla_xJ$ denotes the gradient of the original objective function at the input layer.

The proposed method will add an ''adversarial gradient'' to the learning defined as the gradient difference between adversarial and standard back-propagation:

\begin{equation}
\label{eq:3}
\nabla G = \nabla F(\theta, x, y) - \nabla J(\theta, x, y)
\end{equation}

The adversarial gradient will not vanish unless $\nabla_x J(\theta, x, y)$, becomes exactly zero. This will ensure that adversarial learning continues as the objective function, $J(\theta, x, y)$, approaches zero. The perturbation magnitude $\epsilon$ decides the amount of adversarial training. If the magnitude is zero, the adversarial gradient will vanish, and the method becomes the standard back-propagation.

\begin{algorithm}
\caption{Adversarial back-propagation}
\label{algo}
Repeat for each mini-batch in the training set:
\begin{enumerate}
	\item Propagate the input $x$ forward to the output layer as in standard back-propagation.
	\item Calculate the error and back-propagate the gradient all the way to the input layer.
	\item Calculate the perturbated input $z=x+\epsilon \ sign(e)$, where $e$ is the back-propagated gradient at the input layer from step 2.
	\item Perform forward and backward pass as in standard back-propagation, but use $z$ instead of $x$ as input.
	\item Update weights and biases based on gradients from step 4. 
\end{enumerate}
\end{algorithm}

\section{Experimental results on MNIST}
The MNIST data set is a collection of handwritten digit images. The task is to classify these into 10 classes. The training set consists of 60000 images, and the test set of 10000 images. Here the permutation invariant version of this task is considered.

Some previous result on this task are listed in Table \ref{table:mnist-prev}. The number of published results on this data set is huge, and only a few key results are included here.

\begin{table}[ht]
\caption{Previously published MNIST key results}
\label{table:mnist-prev}
\begin{center}
\begin{tabular}{l|l|l|l}
\multicolumn{1}{l}{\bf METHOD}  &\multicolumn{1}{l}{\bf ARCH} &\multicolumn{1}{l}{\bf UNITS}  &\multicolumn{1}{l}{\bf ERROR}
\\ \hline \\
Back-propagation (\citet{SimardSP03})										&2x800				&logistic         &1.60\% \\
Back-propagation (\citet{WanZZLF13})										&2x800				&tanh         		&1.49\% \\
Back-propagation (\citet{WanZZLF13})										&2x800				&ReLU         		&1.40\% \\
Dropout (\citet{SrivastavaHKSS14})											&3x1024				&logistic					&1.35\% \\
Dropout (\citet{SrivastavaHKSS14})											&3x1024				&ReLU							&1.25\% \\
DropConnect (\citet{WanZZLF13})													&2x800				&ReLU							&1.20\% \\
Dropout + max norm (\citet{SrivastavaHKSS14})						&2x8192				&ReLU							&0.95\% \\
DBM pre-training + dropout (\citet{SrivastavaHKSS14})		&500-1000		  &logistic					&0.79\% \\
Dropout + adversarial examples (\citet{GoodfellowSS14})	&2x1600				&maxout						&0.78\% \\
Virtual Adversarial Training (\citet{MiyatoMKNI15})			&1200-600-300-150	&ReLU					&0.64\% \\
Ladder network (\citet{RasmusVHBR15})										&1000-500-250-250-250	&ReLU			&0.61\% \\
\end{tabular}
\end{center}
\end{table}

\begin{table}[ht]
\caption{Adversarial back-propagation MNIST results}
\label{table:mnist-this}
\begin{center}
\begin{tabular}{l|l|l}
\multicolumn{1}{l}{\bf ARCH} &\multicolumn{1}{l}{\bf UNITS} &\multicolumn{1}{l}{\bf ERROR}
\\ \hline \\
2x400				&logistic   &$1.15\pm0.08\%$\\
2x800				&logistic		&$1.00\pm0.05\%$\\
2x1200			&logistic		&$0.95\pm0.03\%$\\
\\ \hline \\
2x400				&tanh				&$1.04\pm0.04\%$\\
2x800				&tanh				&$1.01\pm0.02\%$\\
2x1200			&tanh				&$1.07\pm0.05\%$\\
\\ \hline \\
2x400				&ReLU				&$0.83\pm0.04\%$\\
2x800				&ReLU				&$0.78\pm0.03\%$\\
2x1200			&ReLU				&$0.78\pm0.03\%$\\
\end{tabular}
\end{center}
\end{table}

Table \ref{table:mnist-this} lists several feed-forward networks trained with adversarial back-propagation. The test set was used for initial experiments. Later, a validation set was used to improve the hyper-parameters, but no extensive search was performed. The input was scaled to be between 0 and 1. The perturbation magnitude $\epsilon=0.08$ was kept constant for all networks. The mini-batch size was 10, and the learning rate was $\alpha=[0.5, 0.01, 0.05]$ for logistic, tanh and ReLU networks. The learning rate was averaged over the number of samples in each mini-batch. The samples were shuffled after each epoch. The last layer was a logistic layer for all networks, and a cross-entropy objective function was used. Weights were initialized to random values drawn from a zero-mean normal distribution with standard deviation 0.01. Biases were initialized to zero. Based on the validation set, all networks were trained for 150 epochs. After this point, error rates did not seem to change much. The error was calculated as the mean of the 10 last epochs averaged over 5 runs. No early stopping, momentum, learning rate schedule, weight decay or weight normalization was used.

\begin{figure}[ht]
	\centering
	\includegraphics[height=0.5\linewidth]{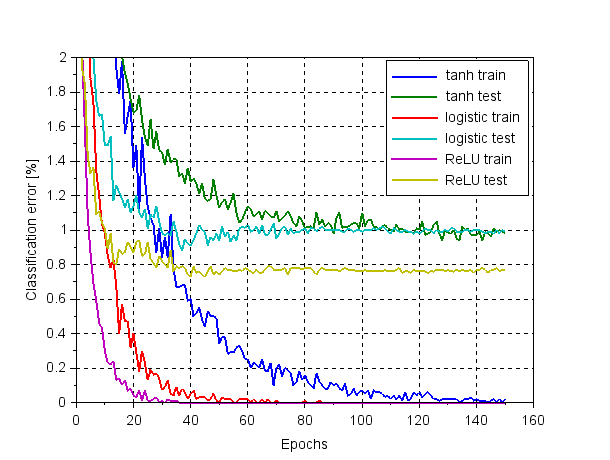}
	\caption{Error curves for 2x800 networks with different activation functions.}
	\label{fig:ErrorCurves}
\end{figure}

Figure (\ref{fig:ErrorCurves}) shows that the error on the clean training set converge to zero even if the networks are trained on perturbated samples only. The figure also shows that the ReLU and logistic networks converge in about 50 epochs, but the tanh network needs more time.

To see if the adversarial training increases the resistance against adversarial examples, a set of adversarial examples was created based on the validation set and the fast gradient sign method with $\epsilon=0.25$. The model used to create the samples was a 2x400 ReLU network trained with standard back-propagation. Another 2x400 ReLU model that was trained with standard back-propagation, classified 22\% of these samples incorrectly. If the same model was trained with adversarial back-propagation, the error decreased to 4\%. For $\epsilon=0.50$, the error decreased from 47\% to 18\%. This means that adversarial back-propagation increases the resistance against adversarial examples generated with the fast gradient sign method.

\begin{figure}[ht]
	\centering
	\includegraphics[height=0.35\linewidth]{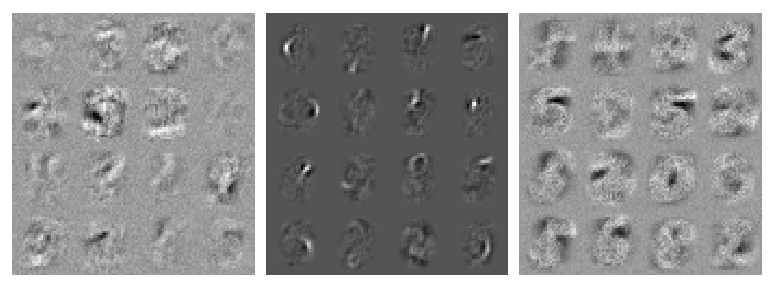}
	\caption{ReLU 1st hidden layer features. (left: back-prop, middle: adversarial back-prop, right: dropout)}
	\label{fig:ReLUFeatures}
\end{figure}

The effect of the adversarial training is apparent when looking at the features of the first hidden layer in a 2x400 ReLU network, see figure (\ref{fig:ReLUFeatures}). The filters are more localized and look more like pen strokes than those obtained with back-propagation with and without dropout.

\begin{figure}[ht]
	\centering
	\includegraphics[height=0.35\linewidth]{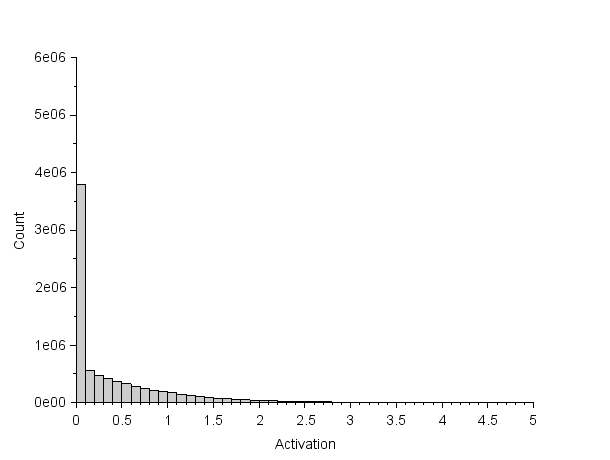}
	\includegraphics[height=0.35\linewidth]{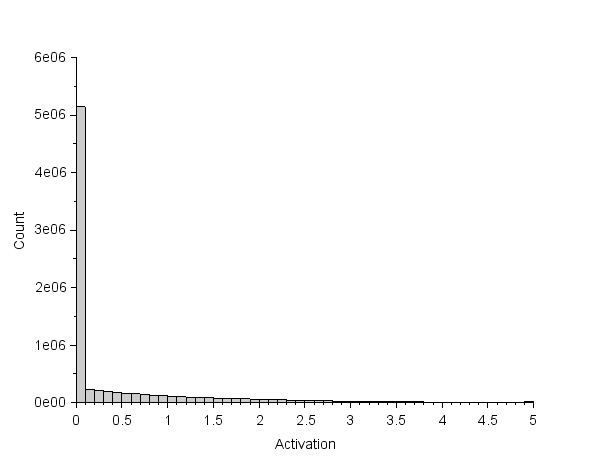}
	\caption{ReLU hidden unit activations. (left: back-prop, right: adversarial back-prop)}
	\label{fig:ReLUActivations}
\end{figure}

Figure (\ref{fig:ReLUActivations}) shows how the adversarial training affects the sparseness of the hidden units in a 2x400 ReLU network. The fraction of real zeros increased from 39\% to 61\% when the network was trained with adversarial back-propagation.

\section{Experimental results on CIFAR-10}
The CIFAR-10 data set, (\citet{Krizhevsky09}), is a collection of color images of dimension 32x32 pixels. The task is to classify these into 10 categorical classes. The training set consists of 50000 images, and the test set of 10000 images. The data was pre-processed in two ways for these experiments. One set of experiments used ZCA whitening, the other set used a simple mean and standard deviation normalization over the whole data set. This was done to investigate how the pre-processing step influences the performance of both dropout and adversarial training.

\begin{table}[ht]
\caption{CIFAR-10 test error for different methods and data pre-processing}
\label{table:cifar-this}
\begin{center}
\begin{tabular}{l|l|l|l}
\multicolumn{1}{l}{\bf ARCH} &\multicolumn{1}{l}{\bf METHOD} &\multicolumn{1}{l}{\bf ZCA} &\multicolumn{1}{l}{\bf SIMPLE}
\\ \hline \\
2x3000			&Back-prop			&$62.72\pm0.29\%$	&$40.57\pm0.33\%$\\
2x3000			&Back-prop + dropout	&$55.45\pm0.22\%$	&$39.51\pm0.31\%$\\
2x3000			&Adversarial			&$55.87\pm0.55\%$	&$38.87\pm0.18\%$\\
2x3000			&Adversarial + dropout	&$52.79\pm0.66\%$&$39.00\pm0.35\%$\\
\\ \hline \\
CONV				&Back-prop			&$14.50\pm0.11\%$	&$15.65\pm0.15\%$\\
CONV				&Back-prop + dropout	&$12.97\pm0.23\%$	&$15.95\pm0.27\%$\\
CONV				&Adversarial			&$13.62\pm0.14\%$	&$17.09\pm0.30\%$\\
CONV				&Adversarial + dropout	&$12.55\pm0.20\%$&$16.00\pm0.46\%$\\
\end{tabular}
\end{center}
\end{table}

Table \ref{table:cifar-this} lists 2 ReLU networks trained on this data set, one fully connected and one convolutional network. 

The convolutional network had 3 convolutional layers and 2 fully connected layers with 2048 neurons each, and was identical to the network used in (\citet{SrivastavaHKSS14}). The weight decay of 0.001, the max-norm constraint of 4, the momentum of 0.9 and the mini-batch size of 128 was not optimized, but taken from this work. The initial learning rate was set to 0.5.

The fully connected network had 2 hidden layers of 3000 neurons each. The max-norm constraint and mini-batch size were as for the convolutional network. No momentum or weight decay was used. The learning rate was set initially to 0.1.

The adversarial perturbation magnitude was increased gradually until the best result on the validation set was achieved. The perturbation magnitude was 0.03 for the fully connected network and 0.01 for the convolutional network. The dropout rate, when used, was 0.1 for the input layer, 0.25 for convolutional layers and 0.5 for fully connected layers.

The last layer was a softmax layer and a cross-entropy objective function was used for both networks. Weights were initialized to random values drawn from a zero-mean normal distribution with standard deviation 0.01. Biases were initialized to zero.

The learning rate was multiplied by 0.5 every time the training error stopped to improve. The training was stopped when the learning rate had been halved 8 times, the training error reached 0.01\%, or the number of epochs reached 300. The test error was calculated as the average over 5 runs.

For the convolutional network and whitened data, the results with and without dropout are similar to those reported in (\citet{SrivastavaHKSS14}). With adversarial training, the result is better than for back-propagation alone, but still not as good as for dropout. The best result is achieved if both methods are combined.

For the fully connected network, adversarial training combined with the simple pre-processing step gives the best result.


\section{Discussion}
The MNIST experiments suggest that training on adversarial examples can replace training on clean samples. This improves the generalization property of feed-forward networks when compared to standard back-propagation. The best result is superior to results where dropout is the only regularizing method. Interestingly, among the three methods that perform equal or better than adversarial back-propagation, all use batch normalization and/or adversarial training.

The 2x400 ReLU network performs better than a 2x8192 network trained with dropout back-propagation. The additional forward and backward pass adds roughly 70\% to the epoch computation time, but the small number of hidden units suggests that the proposed method requires less training time to achieve equal performance.

The CIFAR-10 experiments suggest that adversarial training has a regularizing effect similar to dropout for fully connected networks. For convolutional networks, the effect of both dropout and adversarial back-propagation depends heavily on the kind of pre-processing performed. 

No randomness is introduced in the learning except for initial values and stochastic shuffling of samples in the mini batches. This is in contradiction to pre-training that introduces randomness through sampling or noise, dropout that use randomness for the dropout mask, and noise-injection that, by definition, uses randomness. This means that the learning is less dependent on a proper random number generator.

The question is why adversarial training improves generalization. With adversarial back-propagation, the model will learn to classify samples that are harder to classify that the original ones. We may speculate if this can increase the margin against making errors when used to classify unseen samples, in a similar way as in Support Vector Machines (\citet{BoserGV92}). 

The objective of supervised learning is to discriminate between classes by adjusting hyperplanes or boundaries in a multidimensional input space. For simplicity, assume that the objective is to discriminate between two classes, $A$ and $B$. The boundary between these classes can be defined as the hyperplane where the output units for the two classes have equal values; $y(A)=y(B)$. When trained with standard back-propagation on an input sample from class $A$, the model will try to adjust the boundary in such a way that the sample is placed on the correct side. This is done by increasing $y(A)$ and decreasing $y(B)$ for the input sample. If trained with adversarial back-propagation, the model will instead try to adjust the boundary such that the perturbated input $(x+\epsilon \mu)$ is placed on the correct side. The perturbation has an approximate direction towards the closest boundary. If the model succeeds in placing the perturbated sample on the correct side of the boundary, it will also push the boundary away from the clean data sample. The next time the sample is seen, the closest boundary may be in another direction, and iteratively, the model will increase the margin in all required directions, until balance is achieved.

From this point of view, the reason why adversarial examples are fooling neural networks, is because the margins around the training samples are too small. A boundary may lie indefinitely close to a training sample even if the sample is classified correctly. Even if the loss for this sample is zero, there is no guarantee that nearby points will have zero or small loss. This leads to a connection to the idea of model smoothness. In \citet{MiyatoMKNI15} it is argued that adversarial training increases the smoothness of the model in the neighborhood of the training samples. If the loss function is small for the training sample and at the same time smooth in the vicinity of the sample, this will implicate a good margin.

As stated in \citet{GoodfellowSS14} and \citet{FawziFF15}, adding adversarial perturbations is quite different from adding input noise. Adding noise will direct the model to increase the margin in all possible directions around the training samples. A model has limited capacity, and this may limit the achievable margin in the directions that matters most, where the margins are smallest. 

\section{Conclusion}

The purpose of this paper is to show through preliminary experiments on MNIST and CIFAR-10 that adversarial back-propagation increases the robustness against adversarial examples, and improves generalization in networks with logistic, hyperbolic tangent and rectified linear activation functions. The method performs better than dropout back-propagation on MNIST and is less expensive when it comes to training time, even though an additional forward and backward pass i required. Adversarial back-propagation should be easy to implement in software libraries that already perform back-propagation. 

Further experiments have to be performed to see if the promising results extend to more difficult data sets.

\bibliography{adversarial_backprop}

\begin{thebibliography}{18}
\providecommand{\natexlab}[1]{#1}
\providecommand{\url}[1]{\texttt{#1}}
\expandafter\ifx\csname urlstyle\endcsname\relax
  \providecommand{\doi}[1]{doi: #1}\else
  \providecommand{\doi}{doi: \begingroup \urlstyle{rm}\Url}\fi

\bibitem[Boser et~al.(1992)Boser, Guyon, and Vapnik]{BoserGV92}
Boser, Bernhard~E., Guyon, Isabelle, and Vapnik, Vladimir.
\newblock A training algorithm for optimal margin classifiers.
\newblock In Haussler, David (ed.), \emph{Proceedings of the Fifth Annual {ACM}
  Conference on Computational Learning Theory, {COLT} 1992, Pittsburgh, PA,
  USA, July 27-29, 1992.}, pp.\  144--152. {ACM}, 1992.
\newblock \doi{10.1145/130385.130401}.
\newblock URL \url{http://doi.acm.org/10.1145/130385.130401}.

\bibitem[D.~E.~Rumelhart(1986)]{Hinton86}
D.~E.~Rumelhart, G. E.~Hinton, R. J.~Williams.
\newblock Learning internal representations by error propagation.
\newblock \emph{Nature}, 323:\penalty0 533--536, 1986.

\bibitem[Fawzi et~al.(2015)Fawzi, Fawzi, and Frossard]{FawziFF15}
Fawzi, Alhussein, Fawzi, Omar, and Frossard, Pascal.
\newblock Analysis of classifiers' robustness to adversarial perturbations.
\newblock \emph{CoRR}, abs/1502.02590, 2015.
\newblock URL \url{http://arxiv.org/abs/1502.02590}.

\bibitem[Goodfellow et~al.(2014)Goodfellow, Shlens, and
  Szegedy]{GoodfellowSS14}
Goodfellow, Ian~J., Shlens, Jonathon, and Szegedy, Christian.
\newblock Explaining and harnessing adversarial examples.
\newblock \emph{CoRR}, abs/1412.6572, 2014.
\newblock URL \url{http://arxiv.org/abs/1412.6572}.

\bibitem[Gu \& Rigazio(2014)Gu and Rigazio]{GuR14}
Gu, Shixiang and Rigazio, Luca.
\newblock Towards deep neural network architectures robust to adversarial
  examples.
\newblock \emph{CoRR}, abs/1412.5068, 2014.
\newblock URL \url{http://arxiv.org/abs/1412.5068}.

\bibitem[Hinton et~al.(2006)Hinton, Osindero, and Teh]{HintonOT06}
Hinton, Geoffrey~E., Osindero, Simon, and Teh, Yee~Whye.
\newblock A fast learning algorithm for deep belief nets.
\newblock \emph{Neural Computation}, 18\penalty0 (7):\penalty0 1527--1554,
  2006.
\newblock \doi{10.1162/neco.2006.18.7.1527}.
\newblock URL \url{http://dx.doi.org/10.1162/neco.2006.18.7.1527}.

\bibitem[Ioffe \& Szegedy(2015)Ioffe and Szegedy]{IoffeS15}
Ioffe, Sergey and Szegedy, Christian.
\newblock Batch normalization: Accelerating deep network training by reducing
  internal covariate shift.
\newblock In Bach, Francis~R. and Blei, David~M. (eds.), \emph{Proceedings of
  the 32nd International Conference on Machine Learning, {ICML} 2015, Lille,
  France, 6-11 July 2015}, volume~37 of \emph{{JMLR} Proceedings}, pp.\
  448--456. JMLR.org, 2015.
\newblock URL \url{http://jmlr.org/proceedings/papers/v37/ioffe15.html}.

\bibitem[Krizhevsky \& Hinton(2009)Krizhevsky and Hinton]{Krizhevsky09}
Krizhevsky, Alex and Hinton, Geoffrey.
\newblock Learning multiple layers of features from tiny images.
\newblock \emph{Technical report, University of Toronto}, 2009.
\newblock URL
  \url{https://www.cs.toronto.edu/~kriz/learning-features-2009-TR.pdf}.

\bibitem[Matsuoka(1992)]{Matsuoka92}
Matsuoka, Kiyotoshi.
\newblock Noise injection into inputs in back-propagation learning.
\newblock \emph{{IEEE} Transactions on Systems, Man, and Cybernetics},
  22\penalty0 (3):\penalty0 436--440, 1992.
\newblock \doi{10.1109/21.155944}.
\newblock URL \url{http://dx.doi.org/10.1109/21.155944}.

\bibitem[Miyato et~al.(2015)Miyato, Maeda, Koyama, Nakae, and
  Ishii]{MiyatoMKNI15}
Miyato, Takeru, Maeda, Shin{-}ichi, Koyama, Masanori, Nakae, Ken, and Ishii,
  Shin.
\newblock Distributional smoothing by virtual adversarial examples.
\newblock \emph{CoRR}, abs/1507.00677, 2015.
\newblock URL \url{http://arxiv.org/abs/1507.00677}.

\bibitem[Nguyen et~al.(2014)Nguyen, Yosinski, and Clune]{NguyenYC14}
Nguyen, Anh~Mai, Yosinski, Jason, and Clune, Jeff.
\newblock Deep neural networks are easily fooled: High confidence predictions
  for unrecognizable images.
\newblock \emph{CoRR}, abs/1412.1897, 2014.
\newblock URL \url{http://arxiv.org/abs/1412.1897}.

\bibitem[Rasmus et~al.(2015)Rasmus, Valpola, Honkala, Berglund, and
  Raiko]{RasmusVHBR15}
Rasmus, Antti, Valpola, Harri, Honkala, Mikko, Berglund, Mathias, and Raiko,
  Tapani.
\newblock Semi-supervised learning with ladder network.
\newblock \emph{CoRR}, abs/1507.02672, 2015.
\newblock URL \url{http://arxiv.org/abs/1507.02672}.

\bibitem[Salakhutdinov \& Hinton(2009)Salakhutdinov and
  Hinton]{SalakhutdinovH09}
Salakhutdinov, Ruslan and Hinton, Geoffrey~E.
\newblock Deep boltzmann machines.
\newblock In Dyk, David A.~Van and Welling, Max (eds.), \emph{Proceedings of
  the Twelfth International Conference on Artificial Intelligence and
  Statistics, {AISTATS} 2009, Clearwater Beach, Florida, USA, April 16-18,
  2009}, volume~5 of \emph{{JMLR} Proceedings}, pp.\  448--455. JMLR.org, 2009.
\newblock URL
  \url{http://www.jmlr.org/proceedings/papers/v5/salakhutdinov09a.html}.

\bibitem[Simard et~al.(2003)Simard, Steinkraus, and Platt]{SimardSP03}
Simard, Patrice~Y., Steinkraus, David, and Platt, John~C.
\newblock Best practices for convolutional neural networks applied to visual
  document analysis.
\newblock In \emph{7th International Conference on Document Analysis and
  Recognition {(ICDAR} 2003), 2-Volume Set, 3-6 August 2003, Edinburgh,
  Scotland, {UK}}, pp.\  958--962. {IEEE} Computer Society, 2003.
\newblock \doi{10.1109/ICDAR.2003.1227801}.
\newblock URL \url{http://dx.doi.org/10.1109/ICDAR.2003.1227801}.

\bibitem[Srivastava et~al.(2014)Srivastava, Hinton, Krizhevsky, Sutskever, and
  Salakhutdinov]{SrivastavaHKSS14}
Srivastava, Nitish, Hinton, Geoffrey~E., Krizhevsky, Alex, Sutskever, Ilya, and
  Salakhutdinov, Ruslan.
\newblock Dropout: a simple way to prevent neural networks from overfitting.
\newblock \emph{Journal of Machine Learning Research}, 15\penalty0
  (1):\penalty0 1929--1958, 2014.
\newblock URL \url{http://dl.acm.org/citation.cfm?id=2670313}.

\bibitem[Szegedy et~al.(2013)Szegedy, Zaremba, Sutskever, Bruna, Erhan,
  Goodfellow, and Fergus]{SzegedyZSBEGF13}
Szegedy, Christian, Zaremba, Wojciech, Sutskever, Ilya, Bruna, Joan, Erhan,
  Dumitru, Goodfellow, Ian~J., and Fergus, Rob.
\newblock Intriguing properties of neural networks.
\newblock \emph{CoRR}, abs/1312.6199, 2013.
\newblock URL \url{http://arxiv.org/abs/1312.6199}.

\bibitem[Vincent et~al.(2008)Vincent, Larochelle, Bengio, and
  Manzagol]{VincentLBM08}
Vincent, Pascal, Larochelle, Hugo, Bengio, Yoshua, and Manzagol,
  Pierre{-}Antoine.
\newblock Extracting and composing robust features with denoising autoencoders.
\newblock In \emph{Machine Learning, Proceedings of the Twenty-Fifth
  International Conference {(ICML} 2008), Helsinki, Finland, June 5-9, 2008},
  pp.\  1096--1103, 2008.
\newblock \doi{10.1145/1390156.1390294}.
\newblock URL \url{http://doi.acm.org/10.1145/1390156.1390294}.

\bibitem[Wan et~al.(2013)Wan, Zeiler, Zhang, LeCun, and Fergus]{WanZZLF13}
Wan, Li, Zeiler, Matthew~D., Zhang, Sixin, LeCun, Yann, and Fergus, Rob.
\newblock Regularization of neural networks using dropconnect.
\newblock In \emph{Proceedings of the 30th International Conference on Machine
  Learning, {ICML} 2013, Atlanta, GA, USA, 16-21 June 2013}, volume~28 of
  \emph{{JMLR} Proceedings}, pp.\  1058--1066. JMLR.org, 2013.
\newblock URL \url{http://jmlr.org/proceedings/papers/v28/wan13.html}.

\end{thebibliography}
\bibliographystyle{iclr2016_conference}

\end{document}